\documentclass{article}

\PassOptionsToPackage{square,sort,comma,numbers}{natbib}
\usepackage[final]{nips_2017}
\usepackage{amsmath,bm}
\usepackage{multirow}
\usepackage{diagbox}
\usepackage{booktabs}
\usepackage{arydshln}
\usepackage{graphicx}
\usepackage{algorithm}
\usepackage{multirow}
\usepackage{enumitem}

\usepackage[utf8]{inputenc} 
\usepackage[T1]{fontenc}    
\usepackage{hyperref}       
\usepackage{url}            
\usepackage{booktabs}       
\usepackage{amsfonts}       
\usepackage{nicefrac}       
\usepackage{microtype}      

\usepackage{color}

\def\bm{\boldsymbol}
\def\x{\boldsymbol{x}}
\def\X{\boldsymbol{X}}
\def\p{\boldsymbol{p}}

\title{PatchShuffle Regularization}

\author{
  Guoliang Kang, Xuanyi Dong, Liang Zheng, Yi Yang \\
  Center of AI, University of Technology Sydney, Australia \\
  \textit{\{Guoliang.Kang@student., Xuanyi.Dong@student., Liang.Zheng@, Yi.Yang@\}uts.edu.au}
}

\begin{document}

\maketitle

\begin{abstract}
This paper focuses on regularizing the training of the convolutional neural network (CNN). 
We propose a new regularization approach named ``PatchShuffle'' that can be adopted in any classification-oriented CNN models. It is easy to implement: in each mini-batch, images or feature maps are randomly chosen to undergo a transformation such that pixels within each local patch are shuffled. 
  Through generating images and feature maps with interior orderless patches, PatchShuffle creates rich local variations, reduces the risk of network overfitting, and can be viewed as a beneficial supplement to various kinds of training regularization techniques, such as weight decay, model ensemble and dropout.
  Experiments on four representative classification datasets show that PatchShuffle improves the 
  generalization ability of CNN especially when the data is 
  scarce. Moreover, we empirically illustrate that CNN models trained with PatchShuffle
  are more robust to noise and local changes in an image.

\end{abstract}

\section{Introduction}

The trend of the architectures of deep convolutional neural networks (CNNs) is to become wider \cite{xie2016aggregated,DBLP:conf/bmvc/ZagoruykoK16} and deeper \cite{simonyan2015very,szegedy2015going,he2016deep,he2016identity}. However, millions of parameters make CNNs prone to overfitting when training data is not sufficient. 
In practice, plenty of regularization approaches have been adopted to improve the generalization ability of CNNs, such as weight decay \cite{krogh1991simple}, 
dropout \cite{hinton2012improving,srivastava2014dropout}, batch normalization \cite{ioffe2015batch}, \emph{etc}. This paper provides an alternative regularization option during CNN training with application in image classification. 

Overfitting is a long-standing issue in the machine learning community. The nature of overfitting is that 
the model adapts to the noise rather than capturing the underlying key factors of variations existing 
in the data \cite{zhang2016understanding}. For image classification, when lacking sufficient training data, the learned model may be 
misled by the irrelevant local information which can be regarded as noise.
Moreover, in most classification tasks, it is the overall structure rather than the detailed local pixels that has a large influence on the performance of a CNN model.
The model should be able to identify the input image correctly if   pixels within local structures change in a way that does not destroy the overall view of an image. From another perspective,
if we consider that human will probably not be confused about the the image content under moderate extent of local blur,
it is expected that a data model such as CNN should behave similarly.

In this paper, we propose a new regularization approach: PatchShuffle, which is a beneficial supplement to existing regularization techniques \cite{krogh1991simple,krizhevsky2012imagenet,srivastava2014dropout,ioffe2015batch}. In the training stage, an image or feature map within a mini-batch is randomly chosen to undergo either of the two actions: 1) keep unchanged, or 2) be transformed in such a way that pixels 
within each patch are shuffled. On the one hand, when applied on the images, the shuffled images have nearly the same global structures
with the original ones but possess rich local variations, which are expected to benefit 
the training of CNNs. On the other hand, when PatchShuffle is applied on the feature maps of the convolutional layers, it can be viewed as implementing model ensemble. In fact, locally shuffling 
the pixels within a patch is equivalent to shuffling the convolutional kernels given unshuffled patches. Thus at each iteration, the model is trained from different kernel instantiations. 
PatchShuffle can also be considered to enable weight sharing within each patch. By shuffling, the pixel instantiation at a specific position of an image can be viewed as being sampled from its neighboring pixels within a patch with equal probability. Therefore, across different iterations, patch pixels with different original locations share the same weight.

One might argue that PatchShuffle is a type of data augmentation technique since new images are generated. However, being applied on \emph{a very small percent} of images/feature maps in a mini-batch, we speculate that PatchShuffle is more of a regularization method than data augmentation \footnote{In \cite{lecun2015deep}, data augmentation is considered as belonging to regularization. We differentiate the two concepts in this paper: data augmentation enlarges the training set to a large extent, while regularization makes more elaborate data changes without noticeable enlarging the data volume.}. We also differentiate PatchShuffle from dropout. The latter samples activations from the hidden units, while PatchShuffle samples from all the possible permutations of pixels within patches and no hidden units are discarded.



In summary, the PatchShuffle regularization has the following merits. 
\begin{itemize}[leftmargin=20pt]
\item An efficient method that costs negligible extra time and memories. It can be easily adopted in a variety of CNN models without changing the learning strategy.
\item A complementary technique to existing regularization approaches. On four representative classification datasets, PatchShuffle further improves the classification accuracy when combined with multiple regularization techniques.
\item Improving the robustness of CNNs to data that is noisy or losses partial information. For example, when adding salt-and-pepper noise to the MNIST dataset, our approach outperforms the baseline by more than 20 percent.
\end{itemize}

\section{Related Work}\label{sec:related_work} 
%
We briefly review several aspects that are closely related to this paper, \emph{i.e.,} data augmentation, regularization, and transformation equivariant and invariant networks.

\textbf{Data augmentation.} The direct strategy against overfitting is to train CNNs on more data. Data augmentation addresses this problem by creating new data from existing data to augment the training set. 
Data augmentation is widely adopted in the training of deep neural networks \cite{Goodfellow-et-al-2016,krizhevsky2012imagenet,he2016deep,gan2015learning,lin2013network}.
An effective way to perform data augmentation is to do various transformations, such as flipping, translation, cropping, \emph{etc}. 
From the perspective of generating more images for training, PatchShuffle shares some properties with data augmentation 
methods. 

\textbf{Regularization.}
Regularization is an effective way to reduce the impact of overfitting. Various types of regularization methods have been proposed \cite{hinton2012improving,ioffe2015batch,krogh1991simple,DBLP:conf/nips/SinghHF16,srivastava2014dropout,xie2016disturblabel}.
PatchShuffle relates to two kinds of regularizations. 
1) Model ensemble. It adopts model averaging in which several 
separately trained models vote on the output given a test sample. The voting procedure is robust to prediction errors made by individual classifiers. 
Many methods implicitly implement model ensemble, such as dropout \cite{hinton2012improving,srivastava2014dropout}, stochastic depth \cite{DBLP:conf/eccv/HuangSLSW16} and swapout \cite{DBLP:conf/nips/SinghHF16}. Architectures are averaged by dropout through randomly discarding a group of hidden units, each of which has different widths of layers. Stochastic depth averages architectures with various depths through randomly skipping layers. Swapout samples from abundant set of architectures
with dropout and stochastic depth as its special case.
2) weight sharing. It forces a set of weights 
to be equal \cite{nowlan1992simplifying} and has been used in the architecture of deep convolutional neural networks \cite{lecun1998gradient}. Networks regularized by weight sharing always have transformation invariant 
properties. For example, through a 
weight sharing framework, Ravanbakhsh \emph{et al.} \cite{ravanbakhsh2016deep} propose the permutation equivariant layer that gains robustness to permutations of the input. PatchShuffle is more of a regularization method because the generated images/feature maps share the global structures 
with the original ones and PatchShuffle is applied on a very small amount of images/feature maps.


\textbf{Transformation equivariant and invariant networks.} PatchShuffle regularization is also related to the family of transformation equivariant and invariant networks. Deep symmetry networks \cite{gens2014deep}
generalize vanilla CNN architecture to model arbitrary symmetry groups. 
In \cite{dieleman2016exploiting}, a series of rotation equivariant operations are proposed, such as cyclic slicing, pooling and rolling. 
Ravanbakhsh \emph{et al.} \cite{ravanbakhsh2016deep} propose a kind of permutation equivariant layer that is robust to the permutations of the inputs through designed weight sharing.
All of the aforementioned neural networks mainly focus on the transformations of the whole images and aim to enable the neural networks to be robust to several specific parametric transformation types. They are problem-driven and not easily 
generalized to other datasets.
Moreover, few investigate and exploit various kinds of transformations in the regularizing of 
deep neural networks in a general sense.

In a recent work,  Shen \emph{et al.} \cite{DBLP:conf/aaai/ShenTST17} propose using patch reordering to achieve the rotation and translation invariance. They divide the feature maps into non-overlapping local patches and
reorder the patches according to the $\ell_1$ or $\ell_2$ norm of 
the activations of the patches. 
Their work is similar to PatchShuffle in that they also break the original arrangement of an image or feature maps during training, but critical differences should be clarified.
1) Shen \emph{et al.} \cite{DBLP:conf/aaai/ShenTST17} reorder the \textit{patches}, while we shuffle the \textit{pixels} within each local patch, which does not destroy the global structure.
2) Shen \emph{et al.} \cite{DBLP:conf/aaai/ShenTST17} perform \textit{ranking} according to specific heuristic rule, while PatchShuffle does the \textit{shuffle} operation randomly. Randomness is proved to be useful to regularize
the training of CNNs by explicitly performing model averaging \cite{krizhevsky2012imagenet,hinton2012improving,srivastava2014dropout,DBLP:conf/eccv/HuangSLSW16,DBLP:conf/nips/SinghHF16}.
3) Shen \emph{et al.} \cite{DBLP:conf/aaai/ShenTST17} concentrate on the rotation and translation invariance of models, whereas we adopt PatchShufle as a regularizer. 


\section{PatchShuffle Regularization}\label{sec:method}
\subsection{PatchShuffle Transformation}
\textbf{Formulations.} Let us consider a matrix $\X$ with the size of $N\times N$ elements. A random switch $r$ controls whether $\X$ needs to be transformed (PatchShuffled). Supposing the random variable $r$ subjects to a Bernoulli distribution $r\sim Bernoulli(\epsilon)$, \emph{i.e.,} $r=1$ with probability $\epsilon$ and $r=0$ with probability $1-\epsilon$, the resulted matrix $\tilde{\X}$ can be 
represented as
\begin{equation}\label{eq:1}
\tilde{\X}=(1-r)\X+rT(\X),
\end{equation}
where $T(\cdot)$ denotes the PatchShuffle transformation.
When $\X$ is partitioned into a block matrix with non-overlapping patches of $n\times n$ elements, \emph{i.e.,} 
\begin{equation}\label{eq:2}
\left(
\begin{array}{c:c:c:c}
\text{$\boldsymbol{x}_{11}$} & \text{$\boldsymbol{x}_{12}$} & \cdots & \text{$\boldsymbol{x}_{1,N/n}$} \\
\hdashline
\text{$\boldsymbol{x}_{11}$} & \text{$\boldsymbol{x}_{12}$} & \cdots & \text{$\boldsymbol{x}_{1,N/n}$} \\
\hdashline
\vdots  &  \vdots & \ddots & \vdots \\
\hdashline 
\text{$\boldsymbol{x}_{N/n,1}$} & \text{$\boldsymbol{x}_{N/n,2}$} & \cdots & \text{$\boldsymbol{x}_{k,N/n}$}
\end{array}
\right),
\end{equation}
the PatchShuffle transformation acts on each patch and can be formulated as follows,
\begin{equation}\label{eq:3}
\tilde{\boldsymbol{x}}_{ij} = \p_{ij} \times \boldsymbol{x}_{ij} \times \p^{'}_{ij},
\end{equation}
where $\boldsymbol{x}_{ij}$ denotes a patch located at the $i$-th row and $j$-th column of the block matrix $\boldsymbol{X}$.
$\p_{ij}$ and $\p^{'}_{ij}$ are permutation matrices. 
Pre-multiplying the patch $\x_{ij}$ with $\p_{ij}$ permutes the rows of $\x_{ij}$, whereas post-multiplying the patch $\x_{ij}$ with $\p^{'}_{ij}$
results in the permutation of the columns of $\x_{ij}$.

\begin{figure*}[t]
\begin{center}
\par
\includegraphics[scale=0.37]{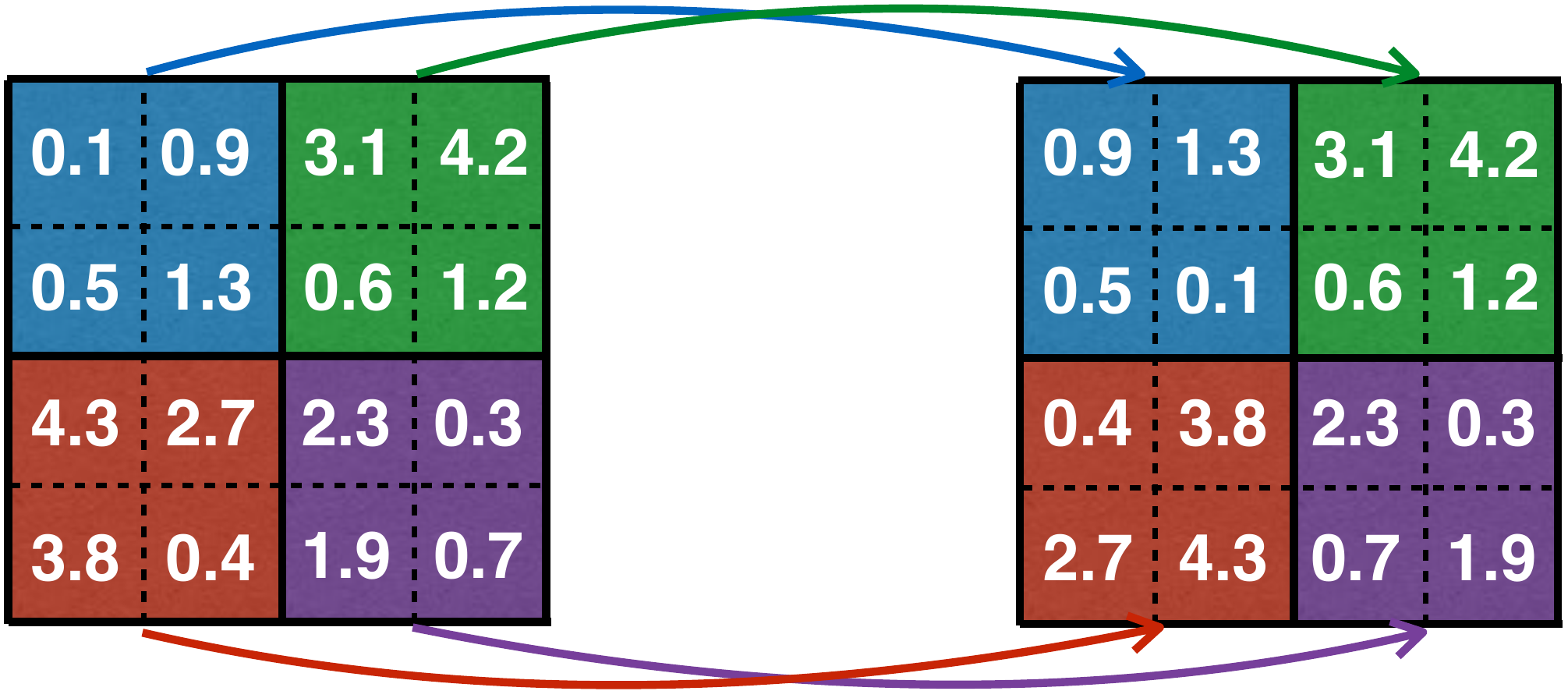}
\end{center}

\caption{\label{fig:schematic-shuffle}{A cartoon illustration  of PatchShuffle on a $4\times 4$ matrix divided into four non-overlapping $2\times 2$ patches
(best viewed in color). Different patches are labeled with different colors.
The shuffling of patches are independent from each other. Note that there is a possibility that pixels within a patch exhibit the original orders after shuffling, as illustrated in the 
upper-right green patch.}}
\end{figure*}

In practice, we first split $\X$ into non-overlapping patches with sizes of $n\times n$ elements. Within
each $n\times n$ patch, the elements are randomly shuffled, as shown in Fig. \ref{fig:schematic-shuffle}. So each patch will undergo one of the $n^2!$ different permutations. For matrix $\X$, the number of possible permutations is $\prod_{i=1}^{N^2/n^2} n^2!$. For each patch, after finite times of shuffle, it will recover the original order. Note that although we describe the PatchShuffle transformation assuming $\X$ and $\x_{ij}$ are square, in practice, $\X$ and its patches $\x_{ij}$ don't need to be square. Our method can be trivially extended to the case of non-square matrix.

\begin{figure*}[t]
\begin{center}
\par
\includegraphics[scale=0.37]{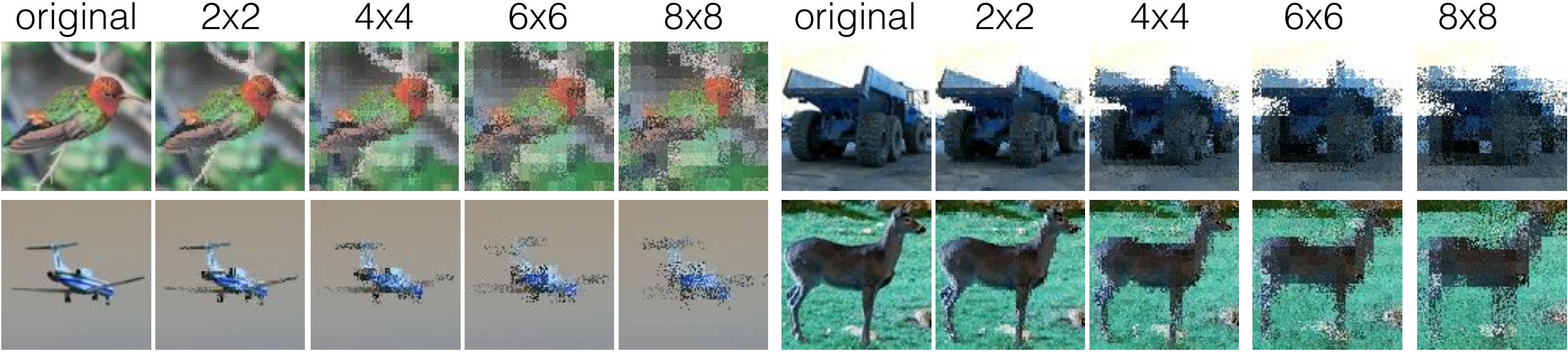}
\end{center}

\caption{\label{fig:shuffle-image}{Samples of PatchShuffled 
real-world images.  The original images are sampled from the STL-10 dataset \cite{coates2011analysis}. The numbers above the  shuffled images denote the patch size $n\times n$ (non-overlapping).}}
\end{figure*}

\textbf{PatchShuffle on images.} 
For CNNs, the input and output of a convolutional layer are feature maps (or images) which can be viewed as matrices. Thus the PatchShuffle transformation is readily applicable.
Following Eq. (\ref{eq:1}), Eq. (\ref{eq:2}) and Eq. (\ref{eq:3}), PatchShuffle can be easily applied on images. Image samples are shown in Fig. \ref{fig:shuffle-image}. Intuitively, we observe that for an image, PatchShuffle transformation within an extent does not disable the recognition of corresponding object, which will benefit the training of a deep neural network. 

\textbf{PatchShuffle on feature maps.}  We also perform PatchShuffle transformation on the feature maps of the \emph{all the convolutional layers}. Here, we treat each feature map as an image, and PatchShuffle is performed on the feature maps independently. That is, each feature map is randomly chosen to undergo the PatchShuffle transformation, regardless of the original image or other feature maps. For the feature maps of lower or middle layers, the spatial structures of the image are preserved to a large extent, so we expect that applying PatchShuffle to these layers can regularize training. For the higher convolutional layers, recall that PatchShuffle enables weight sharing among neighboring pixels; this property is beneficial for the higher level feature maps where neighboring pixels have largely overlapping receptive fields projected onto the original image. We will verify this in the experiment part.


\textbf{Discussions. } PatchShuffle also faces the typical bias-variance dilemma. On the side of reducing the gap between training and test performance,  PatchShuffle creates new  images and feature maps, which increases the variety of the training data. However, on the side of bias, the data distributions of the new images and feature maps created are probably different from those of real-world data, which may induce more bias into the CNN model. Therefore, in the application of PatchShuffle, only a small percentage of the images/feature maps undergo PatchShuffle transformation (small $\epsilon$) to achieve a bias-variance trade-off. 

\subsection{CNN Training and Inference}
\textbf{Objective function.}
We take PatchShuffling images for example. In this case, the training objective function can be 
formalized as,
\begin{equation}
\ell_{s}(\boldsymbol{X},y,\bm{\theta})=(1-r)\ell(\boldsymbol{X},y,\bm{\theta})+r\ell(T(\boldsymbol{X}),y,\bm{\theta}),
\label{eq:objective}
\end{equation}
where $\ell_s$ and $\ell$ denote the objective functions training with and without PatchShuffle, 
respectively. $\bm{X}$ and $T(\bm{X})$ represent the original and PatchShuffled images,
respectively. The label of the training sample is denoted as $y$, and $\bm{\theta}$ encode the weights of the neural network.

In Eq. (\ref{eq:objective}), when the random switch $r$ is set to 1, we have $\ell_{s}(\boldsymbol{X},y,\bm{\theta})=\ell(T(\boldsymbol{X}),y,\bm{\theta})$, which implies 
that network is chosen to be trained with PatchShuffle. When $r=0$, we have $\ell_{s}(\boldsymbol{X},y,\bm{\theta})=\ell(\boldsymbol{X},y,\bm{\theta})$, denoting that the network is trained without PatchShuffle. Taking the expectation over $r$ which follows a Bernoulli distribution, Eq. (\ref{eq:objective})
becomes
\begin{equation}
\frac{1}{1-\epsilon}\mathbb{E}_{r}(\ell_{s}(\boldsymbol{X},y,\bm{\theta)})=\ell(\boldsymbol{X},y,\bm{\theta})+\frac{\epsilon}{1-\epsilon}\ell(T(\boldsymbol{X}),y,\bm{\theta}),
\end{equation}
where $\epsilon$ denotes the shuffle probability, and 
$\frac{\epsilon}{1-\epsilon}\ell(T(\boldsymbol{X}),y,\bm{\theta})$ 
works as a regularizer.

\textbf{Training procedure.} During the training process, PatchShuffle is
applied to the training images and feature maps of all the convolutional layers, each with 
an independently sampled $r$ and independent shuffling operations. Note that the sizes of 
non-overlap patches $H_p\times H_w$ and the shuffle probability ${\epsilon}$ are also not 
necessarily set as the same across layers.
We use the same procedure when applying PatchShuffle to different layers. Without loss of generality,
in Algorithm \ref{alg:patchshuffle}, we summarize the training procedure using PatchShuffle which is applied on the 
the feature maps of one convolutional layer.

\textbf{Inference.} At the test stage, the network performs a forward process \emph{without any transformation} applied to the images or feature maps.
\begin{algorithm}

\caption{PatchShuffle on the feature maps of one convolutional layer in one iteration}

\textbf{Input}: 

1) feature maps $\boldsymbol{X}^{(c)}$ with spatial sizes $H\times W$, $c\in[1,C]$
where $C$ denotes the number of channels of the feature maps in this
convolutional layer.

2) shuffle probability $\epsilon$, patch height $H_{p}$, and patch width
$W_{p}$.

\textbf{Output}: 

feature maps $\tilde{\boldsymbol{X}}^{(c)}$ with the same spatial
sizes as $\boldsymbol{X}^{(c)}$, where $c\in[1,C]$.

Let $\boldsymbol{Z}^{(c)}$ represents the corresponding mapping
between $\tilde{\boldsymbol{X}}^{(c)}$ and $\boldsymbol{X}^{(c)}$. 

\textbf{Forward process}:

$h=\left\lceil H/H_{p}\right\rceil $, $w=\left\lceil W/W_{p}\right\rceil $

for $c$ = 1 to $C$:

$\qquad$Generate the random switch $r$;

$\qquad$if $r=1$:
shuffle each of the $h\times w$ patches of $\boldsymbol{X}^{(c)}$ and storing the mapping $\boldsymbol{Z}^{(c)}$

$\qquad$else:
$\tilde{\boldsymbol{X}}^{(c)}=\boldsymbol{X}^{(c)}$

end     

\textbf{Backward process}:

for $c$ = 1 to $C$:

$\qquad$if $r=1$:
mapping from the gradients of $\tilde{\boldsymbol{X}}^{(c)}$
to those of $\boldsymbol{X}^{(c)}$ according to $\boldsymbol{Z}^{(c)}$.

$\qquad$else:
copying from the gradients of $\tilde{\boldsymbol{X}}^{(c)}$
to those of $\boldsymbol{X}^{(c)}$ 

end
\label{alg:patchshuffle}
\end{algorithm}

%
%
%
%
%
%

\section{Experiments}\label{sec:experiment}
In this section, we report results on four image classification datasets including CIFAR-10, SVHN, STL-10 and MNIST.
CIFAR-10 \cite{krizhevsky2009learning} contains 50,000+10,000 (training+test) 32 $\times$ 32 color images of 10 object classes.
SVHN \cite{netzer2011reading} consists 73,257+26,032 (training+test) $32\times32$ color images for street view house numbers.
STL-10 \cite{coates2011analysis} contains 5,000+8,000 (training+test) $96\times96$ color images for 10 categories.
MNIST \cite{lecun1998gradient} consists of 60,000+10,000 (training+test) 28 $\times$ 28 greyscale images of hand-written digits.
We first show that our algorithm is robust to the change of hyper-parameters within a wide range. Then we demonstrates the improved generalization ability achieved by PatchShuffle on the benchmarks. Finally, we illustrate that CNNs trained by PatchShuffle are more robust to the noises such as salt-and-pepper, occlusions, \emph{etc}. 
All experiments are implemented using \textit{Caffe} \cite{jia2014caffe}.

\subsection{Experiment Settings}
In all of our experiments, we compare the CNN models trained with and without PatchShuffle. 
The training method without PatchShuffle is denoted as standard back-propagation (BP). 
For the same deep architecture, all the models are trained from the same weight initializations.
Note that some popular regularization techniques (\emph{i.e.,} weight decay, batch normalization and dropout) and various data augmentations (\emph{i.e.,} flipping, padding
and cropping) are employed in the experiments. 
The hyper-parameters for training with PatchShuffle and standard BP are all
the same except that the patch size $H_{p}\times W_{p}$ and shuffle probability $\epsilon$ are chosen through validation for
PatchShuffle.
Our experiments build on various CNN architectures, which are summarized as follows:

\textbf{CNNs for CIFAR-10.} Three CNN models are adopted in the experiments of CIFAR-10:
Network in Network \cite{lin2013network} (NIN), pre-activation ResNet-110 \cite{he2016identity} (ResNet-110-PreAct), and the modification of original ResNet-110 \cite{he2016deep} (ResNet-110-Modified). The architecture of ResNet-110-Modified is the same as the original ResNet-110 designed for CIFAR \cite{he2016deep} except that it discards the ReLU unit after each summation of the shortcut and residual function. This small modification improves the performance of original ResNet to be comparable to that of pre-activation ResNet \cite{he2016identity}. The training procedure of ResNet-110-PreAct and ResNet-110-Modified is the same with \cite{he2016deep,he2016identity}, and that of NIN is the same with \cite{lin2013network}.

\textbf{CNNs for SVHN.} 
The CNNs adopted for SVHN are Plain-SVHN and ResNet-110-Modified.
The architecture of Plain-SVHN is the same as provided in the \textit{Caffe} examples\footnote{https://github.com/BVLC/caffe/blob/master/examples/cifar10/cifar10\_full\_train\_test.prototxt} which is originally designed by \cite{krizhevskycuda} for the training of CIFAR-10. Because the architecture is general and the image resolutions are the same between SVHN and CIFAR-10, it can be employed for SVHN. 

\textbf{CNNs for STL-10.} 
The CNN architecture adopted for STL-10 is denoted by ResNet-STL-10.
It is similar to ResNet-110-Modified, but due to a higher resolution of the images, several modifications are made. 1) The kernel size for the first convolutional layer becomes 7 with 
the stride of 2. 2) Four residual stages are employed with each of them containing two residual units.
The spatial sizes of feature maps are successively halved after each residual stage. The channels of the feature maps for four residual stages are 32, 64, 128 and 256, respectively.
3) Dropout is applied on the final fully-connected layer with dropout ratio of 0.5. 

\textbf{CNNs for MNIST.} The CNN architecture adopted on MNIST is provided in \textit{Caffe} examples\footnote{https://github.com/BVLC/caffe/blob/master/examples/mnist/lenet\_train\_test.prototxt}.
\subsection{The Impact of Hyper-parameters}

\begin{figure}[t]
\par
\begin{center}
\includegraphics[bb=0bp 200bp 595bp 642bp,scale=0.30]{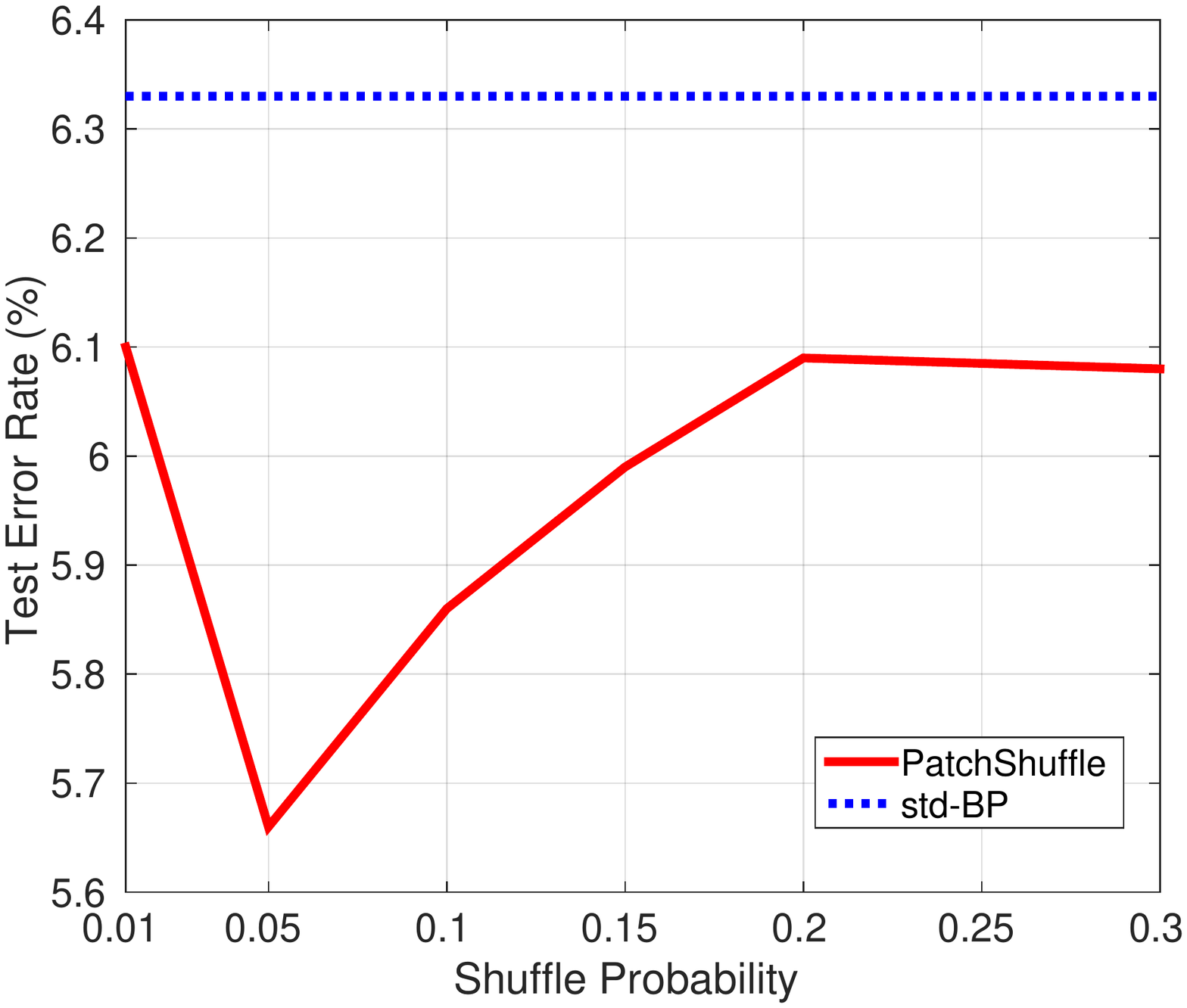}
\qquad
\includegraphics[bb=0bp 200bp 595bp 642bp,scale=0.30]{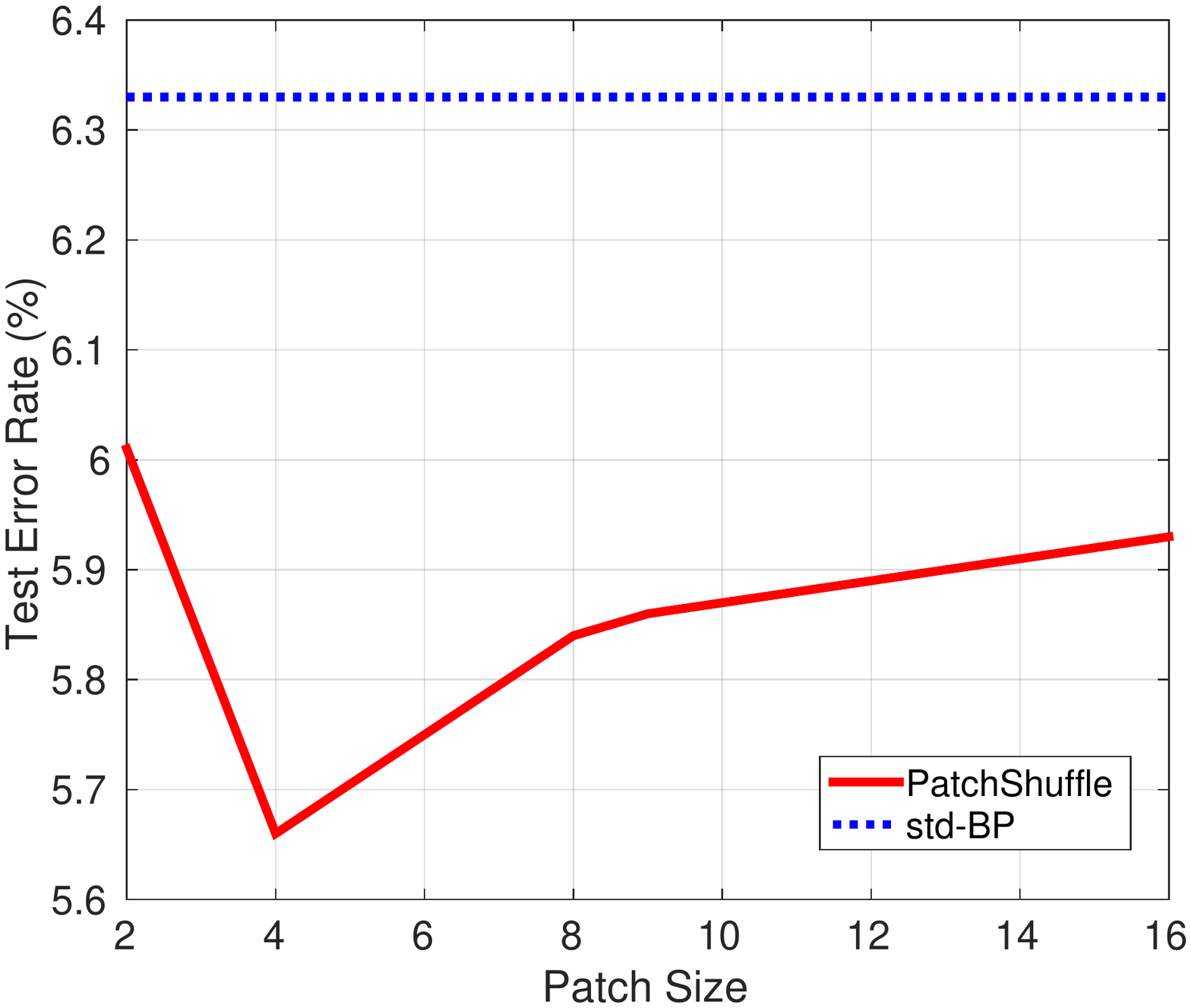}
\end{center}
\caption{\label{fig:hyper-param}{Test error (\%) vs. different hyper-parameters. \textbf{Left}: shuffle probability. \textbf{Right}: patch size. We adopt the minimum value of test errors among different settings of one kind of hyper-parameters to represent the performance of another. }}
\end{figure}

When applying PatchShuffle to CNN training, we have two hyper-parameters to evaluate, \emph{i.e.,}  the patch size $H_p \times W_p$ and the shuffle probability ${\epsilon}$. To demonstrate the impact of these two hyper-parameters on the performance of the model, 
we conduct experiments on CIFAR-10 based on RestNet-110-Modified under different hyper-parameter settings with PatchShuffle applied on the images. The results are compared with standard BP and shown in percentage in Table \ref{table:prob_patch}. Note that all the models are trained with 
the simple data augmentation as in \cite{he2016deep,he2016identity}: 4 pixels are padded on each side, and a 32 $\times$ 32 crop is randomly sampled from the padded image or its horizontal flip. Results are presented in  Table \ref{table:prob_patch} and Fig. \ref{fig:hyper-param}. We arrive at two findings. 

\begin{table}[ht]
\begin{center}
\begin{tabular}{ c | c | c   c   c   c   c | c}
\hline
\diagbox{Probability}{Patch Size}
       & std-BP   &  $1\times2$  &  $2\times2$  &  $2\times4$  &  $3\times3$  &  $4\times4$  & Min. \\\hline
0.01   & \multirow{6}{*}{6.33}      & 6.10     & 6.17 & 6.26 & 6.34 & 6.40 & 6.10 \\
0.05   &         & 6.05     & 5.66 & 5.84 & 5.93 & 6.06 & 5.66 \\
0.10   &         & 6.01     & 5.95 & 6.08 & 5.86 & 5.93 & 5.86 \\
0.15   &         & 6.25     & 6.27 & 5.99 & 6.29 & 6.10 & 5.99 \\
0.20   &         & 6.16     & 6.09 & 6.32 & 6.62 & 6.22 & 6.09 \\
0.30   &         & 6.08     & 6.56 & 6.83 & 7.17 & 6.86 & 6.08 \\\hline
Min.   & 6.33    & 6.01     & 5.66 & 5.84 & 5.86 & 5.93 & \textbf{5.66} \\\hline
\end{tabular}
\end{center}
\caption{Test errors (\%) on CIFAR-10 based on ResNet-110-Modified under different hyper-parameters.}

\label{table:prob_patch}
\end{table}

First, PatchShuffle consistently outperforms standard BP under a wide range of hyper-parameters. On CIFAR-10, our best result reduces the classification error by 0.67\% compared with standard BP. 

Second, PatchShuffle is robust to parameter changes to some extent. When the shuffle probability and patch size increase, recognition error first decreases, touches the bottom, and then increases. In fact, within an extent, the increase of both parameters improve the variety of training sample without introducing too much bias. But under larger values, the benefit brought by diversity is gradually overtaken by the classifier bias, so error rate increases. In the following experiments, we use $\epsilon = 0.05$, patch size = $2\times 2$ when not specified.


\subsection{Classification Performance}

\textbf{CIFAR-10.} 
For the training of NIN and ResNet-110-Modified, we apply PatchShuffle on the images only, while for 
the training of ResNet-110-PreAct, PatchShuffle is applied on the images and feature maps between 
two successive convolutional layers in each residual unit.
The test errors are shown in percentage in Table  \ref{tab:cifar10-full}.
It can be seen that the models trained by PatchShuffle are consistently superior to  those trained by standard BP with using the three CNN architectures.

We further evaluate the impact of the size of training set on recognition accuracy in CIFAR-10. We use the ResNet-110-Modified model. Results are shown in Table  \ref{tab:cifar10-full}. In all of the experiments, we use the same hyper-parameter setting (\emph{i.e.,} with patch size $2\times 2$, and shuffle probability 0.05 ). Although the hyper-parameter setting may not be optimal under small training sets,  Table  \ref{tab:cifar10-full} still indicates that PatchShuffle 
improves the recognition accuracy, especial when the training set is small (see the 
results when training set size is 9,000).

\begin{table}[ht]
\begin{center}
\begin{tabular}{l | c c}
\hline
Model                                   & std-BP    & PatchShuffle    \\\hline
NIN\cite{lin2013network}                & 10.43     &\textbf{10.09}            \\
ResNet-110-PreAct\cite{he2016identity}  & 6.37      & \textbf{5.82}           \\
ResNet-110-Modified                     & 6.33      & \textbf{5.66}           \\\hline
\end{tabular}
\qquad
\begin{tabular}{r | c c}
\hline 
Size                       &  std-BP  & PatchShuffle \\\hline
9,000                      &  22.12   & \textbf{18.20}        \\
15,000                     &  14.26   & \textbf{13.76}        \\
24,000                     &   9.83   & \textbf{9.64}       \\
40,000                     &   7.17   & \textbf{6.66}       \\
50,000                     &   6.33   & \textbf{5.66}       \\\hline
\end{tabular}
\end{center}
\caption{Test errors (\%) on CIFAR-10. \textbf{Left} : different CNN architectures. \textbf{Right} : different sizes of training data. }
\label{tab:cifar10-full}
\end{table}
\textbf{SVHN.} PatchShuffle is applied on the images.
The results are shown in Table  \ref{tab:svhn}, indicating that PatchShuffle consistently outperforms standard BP.

\textbf{STL-10.} Here we illustrate the impact of applying PatchShuffle on the feature maps of CNNs.

\begin{table}[ht]
\begin{center}
\begin{tabular}{l | c c }
\hline
Model                            & std-BP   & PatchShuffle    \\\hline
Plain-SVHN                       & 6.24     & \textbf{5.85}            \\
ResNet-110-Modified              & 4.73     & \textbf{4.11}            \\\hline
\end{tabular}
\end{center}
\caption{Test errors (\%) on SVHN. No data augmentation is adopted. No other data preprocessing is
performed other than per-pixel mean computed over the training set is subtracted from each image.}
\label{tab:svhn}
\end{table}

The classification results on STL-10 are summarized in Table  \ref{tab:stl-10}. The five-bit binary code denotes on what stages PatchShuffle is applied. The first bit denotes the 
input layer, and the other four bits correspond to four residual stages.
Applying PatchShuffle on a stage of ResNet means applying it on the feature maps between two adjacent convolutional layers of each residual unit in this stage.
We set $\epsilon$ to 0.30 in this experiment. 

Table  \ref{tab:stl-10} reveals that the generalization ability of the CNN models trained with PatchShuffle are significantly higher than using standard BP. More significant improvement over the baseline can be observed when using PatchShuffle on more convolutional layers. In addition, increasing the sizes of the patches also brings notable improvement in terms of the generalization performance. Note that all the models are trained \emph{with dropout} applied on the output layer, which suggests that PatchShuffle can reduce overfitting beyond dropout.

\begin{table}[ht]
\begin{center}
\begin{tabular}{c|c|ccccc}
\hline 
\diagbox{Patch Size}{Layers} & std-BP &  10000 & 11000 & 11100 & 11110 & 11111   \\\hline
$2\times2$                    &\multirow{3}{*}{50.42} &47.49   &43.57   &40.66   &36.08   &35.07 \\
$4\times4$                    & &45.85   &41.26   &37.26   &33.75   &\textbf{33.16} \\
$6\times6$                    & &43.04   &40.80   &36.99   &33.45   &33.19 \\\hline
\end{tabular}
\end{center}
\caption{Test errors (\%) on STL-10 with different patch sizes and different choices of layers on which PatchShuffle is applied.}
\label{tab:stl-10}
\end{table}

\begin{table}[ht]
\begin{center}
\begin{tabular}{c|cccc}
\hline 
${\tau_1}$                           & std-BP*  &PS*             & std-BP & PS             \\\hline
0.1                                  & 11.78    & \textbf{2.67 } & 1.44   & \textbf{1.17}   \\ 
0.3                                  & 48.72    & \textbf{24.15} & 3.32   & \textbf{2.67}   \\
0.5                                  & 78.41    & \textbf{54.67} &10.02   & \textbf{8.39}   \\
0.7                                  & 88.88    & \textbf{75.61} &37.34   & \textbf{28.85}  \\
0.9                                  & 90.24    & \textbf{87.37} &83.87   & \textbf{76.26}  \\\bottomrule
\end{tabular}
\quad
\begin{tabular}{c|cc}
\hline 
$\tau_2$                                 & std-BP & PS \\\hline
0.05                                  &4.55    & \textbf{4.40 }       \\ 
0.10                                  &11.40   & \textbf{10.77}         \\
0.15                                  &22.83   & \textbf{20.97}         \\
0.20                                  &36.26   & \textbf{33.12}        \\
0.25                                  &51.04   & \textbf{47.58}         \\\bottomrule
\end{tabular}
\end{center}
\caption{Test errors (\%) on MNIST under different levels of noise and occlusions. 
``PS'' indicates PatchShuffle. All the test sets are polluted with noise or occlusions. 
``*'' means that the training set is not polluted.
None ``*'' denotes that the training set is also polluted with the same level of noise/occlusions with the test set. \textbf{Left}: salt-and-pepper noise. \textbf{Right}: occlusions.
}
\label{tab:noise-sp-1}
\end{table}

\subsection{Robustness to the Noise}
Finally, we show the robustness of PatchShuffle against noise and occlusions. In experiment, we add different levels of salt-and-pepper noise and occlusions on the MNIST dataset. Salt-and-pepper noise is added to the image by changing the pixel to white or black with probability $\tau_1$. For the occlusion, each pixel is randomly chosen to be imposed by
a black block of certain size centered on it with probability $\tau_2$.
The size of the block adopted in our 
experiment is $3 \times 3$.
Results are presented in Table  \ref{tab:noise-sp-1}.


A clear observation is that under increasing level of pollution, the performances of both standard BP and PatchShuffle drop quickly. Nevertheless, under each pollution level, our method yields consistently lower error rate than standard BP. For salt-and-pepper noise, the performance gap is largest (23.74\%) under a noise level of 50\%. For occlusion, our method exceeds standard BP by 3.46\% under occlusion extent of 25\%. These results indicate that Patchshuffle improves the robustness of CNNs against common image pollutions like noise and occlusion.


\section{Conclusion}\label{sec:conclusions}
This paper introduces PatchShuffle, a new regularization method for generalizable CNN training. This method is efficient to compute, complementary to existing regularizers, and improves CNN's robustness to noise and occlusions. We will explore Patchshuffle on more complex tasks in future, \emph{e.g.,} object detection and language modeling. 

\bibliographystyle{plain}
\bibliography{patchshuffle}

\end{document}